%% file: main.tex
\documentclass[runningheads]{llncs}
\usepackage{graphicx}
\usepackage{comment}
\usepackage{amsmath,amssymb} 
\usepackage{color}

\usepackage{siunitx}
\usepackage{cite}
\usepackage{ctable}
\usepackage{hhline}
\usepackage{booktabs}
\usepackage{csquotes}
\usepackage{pifont}
\usepackage{multirow}
\usepackage{booktabs}
\usepackage{hyperref}
\usepackage[super]{nth}

\newcolumntype{L}[1]{>{\raggedright\let\newline\\\arraybackslash\hspace{0pt}}m{#1}}
\newcolumntype{C}[1]{>{\centering\let\newline\\\arraybackslash\hspace{0pt}}m{#1}}
\newcolumntype{R}[1]{>{\raggedleft\let\newline\\\arraybackslash\hspace{0pt}}m{#1}}

\newcommand{\cmark}{\ding{51}}%
\newcommand{\xmark}{\ding{55}}%

\begin{document}
\pagestyle{headings}
\mainmatter

\title{I2L-MeshNet: Image-to-Lixel Prediction Network for Accurate 3D Human Pose and Mesh Estimation from a Single RGB Image} 

\titlerunning{I2L-MeshNet}
%
\author{Gyeongsik Moon \and
Kyoung Mu Lee}
\authorrunning{G. Moon and K. M. Lee}
%
\institute{ECE \& ASRI, Seoul National University, Korea \\
\email{\{mks0601,kyoungmu\}@snu.ac.kr}}
\maketitle

\begin{abstract}
Most of the previous image-based 3D human pose and mesh estimation methods estimate parameters of the human mesh model from an input image.
However, directly regressing the parameters from the input image is a highly non-linear mapping because it breaks the spatial relationship between pixels in the input image. 
In addition, it cannot model the prediction uncertainty, which can make training harder.
To resolve the above issues, we propose I2L-MeshNet, an image-to-lixel (line+pixel) prediction network.
The proposed I2L-MeshNet predicts the per-lixel likelihood on 1D heatmaps for each mesh vertex coordinate instead of directly regressing the parameters.
Our lixel-based 1D heatmap preserves the spatial relationship in the input image and models the prediction uncertainty.
We demonstrate the benefit of the image-to-lixel prediction and show that the proposed I2L-MeshNet outperforms previous methods.
The code is publicly available \footnote{\url{https://github.com/mks0601/I2L-MeshNet_RELEASE}}.
\end{abstract}

\input{src/introduction.tex}
\input{src/related_works.tex}

\input{src/i2l-meshnet.tex}

\input{src/implementation_details.tex}
\input{src/experiment.tex}

\input{src/conclusion.tex}

\section*{Acknowledgments}
This work was supported by IITP grant funded by the Ministry of Science and ICT of Korea (No. 2017-0-01780), and Hyundai Motor Group through HMG-SNU AI Consortium fund (No. 5264-20190101).

\input{src/suppl.tex}

%
%
\bibliographystyle{splncs04}
\bibliography{main}
\end{document}

%% file: src/introduction.tex
\section{Introduction}

\begin{figure}[t]
\begin{center}
\includegraphics[width=1.0\linewidth]{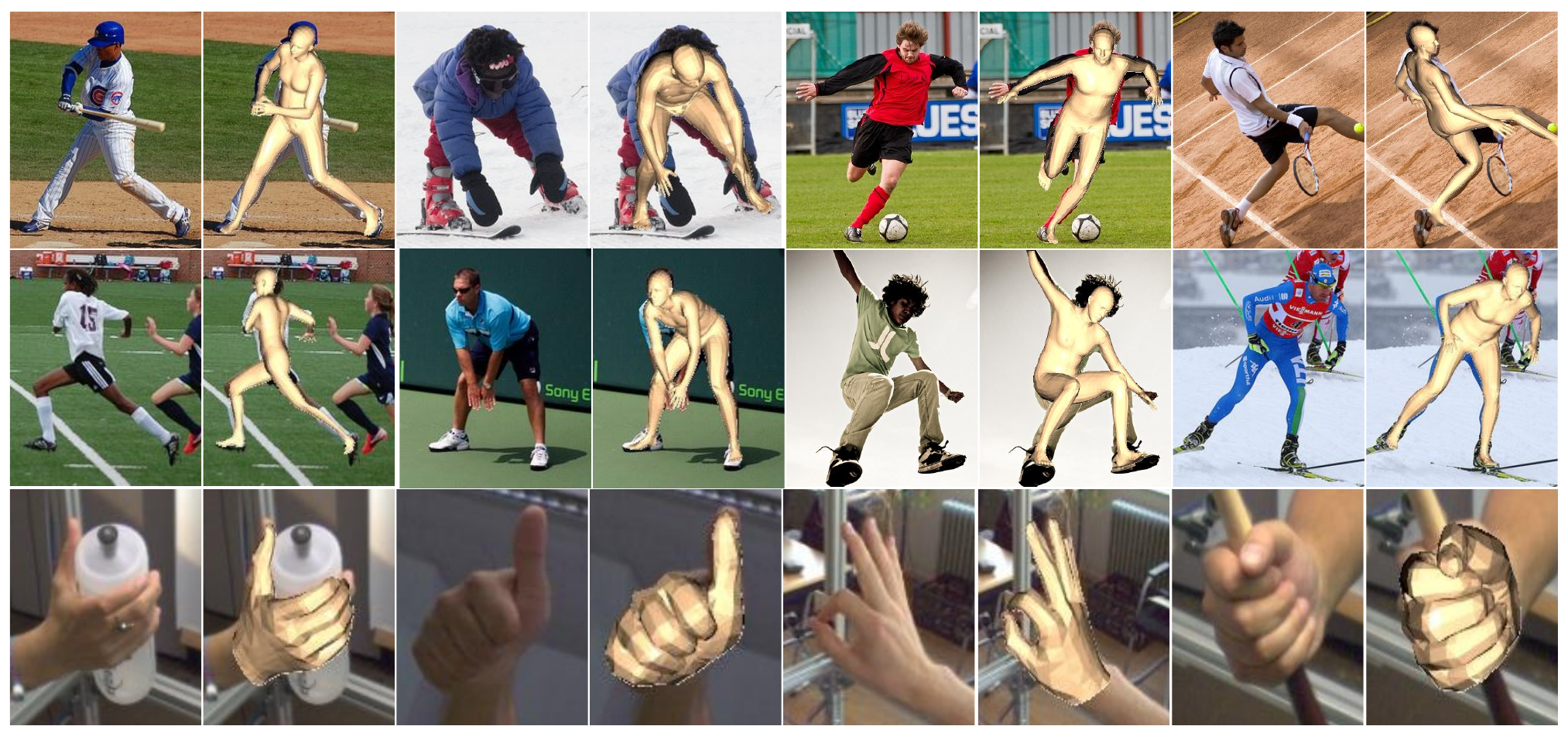}
\end{center}
   \caption{
   Qualitative results of the proposed I2L-MeshNet on MSCOCO~\cite{lin2014microsoft} and FreiHAND~\cite{Freihand2019} datasets.
   }
\label{fig:intro_qualitative}
\end{figure}

3D human pose and mesh estimation aims to simultaneously recover 3D semantic human joint and 3D human mesh vertex locations. 
This is a very challenging task because of complicated human articulation and 2D-to-3D ambiguity.
It can be used in many applications such as virtual/augmented reality and human action recognition.

SMPL~\cite{loper2015smpl} and MANO~\cite{romero2017embodied} are the most widely used parametric human body and hand mesh models, respectively, which can represent various human poses and identities.
They produce 3D human joint and mesh coordinates from pose and identity parameters.
Recent deep convolutional neural network (CNN)-based studies~\cite{kanazawa2018end,pavlakos2018learning,kolotouros2019learning} for the 3D human pose and mesh estimation are based on the model-based approach, which trains a network to estimate SMPL/MANO parameters from an input image. 
On the other hand, there have been few methods based on model-free approach~\cite{ge20193d,kolotouros2019convolutional}, which estimates mesh vertex coordinates directly.
They obtain the 3D pose by multiplying a joint regression matrix, included in the human mesh model, to the estimated mesh.

Although the recent deep CNN-based methods perform impressive, when estimating the target (\textit{i.e.}, SMPL/MANO parameters or mesh vertex coordinates), all of the previous 3D human pose and mesh estimation works break the spatial relationship among pixels in the input image because of the fully-connected layers at the output stage.
In addition, their target representations cannot model the uncertainty of the prediction.
The above limitations can make training harder, and as a result, reduce the test accuracy as addressed in~\cite{moon2018v2v,tompson2014joint}.
To address the limitations, recent state-of-the-art 3D human pose estimation methods~\cite{moon2018v2v,moon2019camera,sun2018integral}, which localize 3D human joint coordinates without mesh vertex coordinates, utilize the \emph{heatmap} as the target representation of their networks.
Each value of one heatmap represents the likelihood of the existence of a human joint at the corresponding pixel positions of the input image and discretized depth value.
Therefore, it preserves the spatial relationship between pixels in the input image and models the prediction uncertainty.

Inspired by the recent state-of-the-art heatmap-based 3D human pose estimation methods, we propose I2L-MeshNet, image-to-lixel prediction network that naturally extends heatmap-based 3D human pose to heatmap-based 3D human pose and mesh.
Likewise voxel (volume+pixel) is defined as a quantized cell in three-dimensional space, we define \emph{lixel (line+pixel)} as a quantized cell in one-dimensional space.
Our I2L-MeshNet estimates per-lixel likelihood on 1D heatmaps for each mesh vertex coordinates, therefore it is based on the model-free approach.
The previous state-of-the-art heatmap-based 3D human pose estimation methods predict 3D heatmap of each human joint.
Unlike the number of human joints, which is around 20, the number of mesh vertex is much larger (\textit{e.g.}, 6980 for SMPL and 776 for MANO).
As a result, predicting 3D heatmaps of all mesh vertices becomes computationally infeasible, which is beyond the limit of modern GPU memory.
In contrast, the proposed lixel-based 1D heatmap has an efficient memory complexity, which has a linear relationship with the heatmap resolution.
Thus, it allows our system to predict heatmaps with sufficient resolution, which is essential for dense mesh vertex localization.

For more accurate 3D human pose and mesh estimation, we design the I2L-MeshNet as a cascaded network architecture, which consists of PoseNet and MeshNet. 
The PoseNet predicts the lixel-based 1D heatmaps of each 3D human joint coordinate.
Then, the MeshNet utilizes the output of the PoseNet as an additional input along with the image feature to predict the lixel-based 1D heatmaps of each 3D human mesh vertex coordinate.
As the locations of the human joints provide coarse but important information about the human mesh vertex locations, utilizing it for 3D mesh estimation is natural and can increase accuracy substantially.

Our I2L-MeshNet outperforms previous 3D human pose and mesh estimation methods on various 3D human pose and mesh benchmark datasets.
Figure~\ref{fig:intro_qualitative} shows 3D human body and hand mesh estimation results on publicly available datasets.

Our contributions can be summarized as follows.
\begin{itemize}
\item We propose I2L-MeshNet, a novel image-to-lixel prediction network for 3D human pose and mesh estimation from a single RGB image.
Our system predicts lixel-based 1D heatmap that preserves the spatial relationship in the input image and models the uncertainty of the prediction.
\item Our efficient lixel-based 1D heatmap allows our system to predict heatmaps with sufficient resolution, which is essential for dense mesh vertex localization.
\item We show that our I2L-MeshNet outperforms previous state-of-the-art methods on various 3D human pose and mesh datasets.
\end{itemize}

%% file: src/related_works.tex
\section{Related works}

\noindent \textbf{3D human body and hand pose and mesh estimation.}
Most of the current 3D human pose and mesh estimation methods are based on the model-based approach, which predict parameters of pre-defined human body and hand mesh models (\textit{i.e.}, SMPL and MANO, respectively).
The model-based methods can be trained only from groundtruth human joint coordinates without mesh vertex coordinates because the model parameters are embedded in low dimensional space.
Early model-based methods~\cite{bogo2016keep} iteratively fit the SMPL parameters to estimated 2D human joint locations. 
More recent model-based methods regress the body model parameters from an input image using CNN. 
Kanazawa~et al.~\cite{kanazawa2018end} proposed an end-to-end trainable human mesh recovery (HMR) system that uses the adversarial loss to make their output human shape is anatomically plausible. 
Pavlakos~et al.~\cite{pavlakos2018learning} used 2D joint heatmaps and silhouette as cues for predicting accurate SMPL parameters. 
Omran~et al.~\cite{omran2018neural} proposed a similar system, which exploits human part segmentation as a cue for regressing SMPL parameters. 
Xu~et al.~\cite{xu2019denserac} used differentiable rendering to supervise human mesh in the 2D image space. 
Pavlakos~et al.~\cite{pavlakos2019texturepose} proposed a system that uses multi-view color consistency to supervise a network using multi-view geometry. 
Baek~et al.~\cite{baek2019pushing} trained their network to estimate the MANO parameters using a differentiable renderer. 
Boukhayma~et al.~\cite{boukhayma20193d} trained their network that takes a single RGB image and estimates MANO parameters by minimizing the distance of the estimated hand joint locations and groundtruth.
Kolotouros~et al.~\cite{kolotouros2019learning} introduced a self-improving system consists of SMPL parameter regressor and iterative fitting framework~\cite{bogo2016keep}.

On the other hand, the model-free approach estimates the mesh vertex coordinates directly instead of regressing the model parameters. 
Due to the recent advancement of the iterative human body and hand model fitting frameworks~\cite{bogo2016keep,pavlakos2019expressive,Freihand2019}, pseudo-groundtruth mesh vertex annotation on large-scale datasets~\cite{ionescu2014human3,lin2014microsoft,Freihand2019,von2018recovering} became available.
Those datasets with mesh vertex annotation motivated several model-free methods that require mesh supervision.
Kolotouros~et al.~\cite{kolotouros2019convolutional} designed a graph convolutional human mesh regression system.
Their graph convolutional network takes a template human mesh in a rest pose as input and outputs mesh vertex coordinates using image feature from ResNet~\cite{he2016deep}.
Ge~et al.~\cite{ge20193d} proposed a graph convolution-based network which directly estimates vertices of hand mesh.
Recently, Choi~et al.~\cite{choi2020p2m} proposed a graph convolutional network that recovers 3D human pose and mesh from a 2D human pose.

Unlike all the above model-based and model-free 3D human pose and mesh estimation methods, the proposed I2L-MeshNet outputs 3D human pose and mesh by preserving the spatial relationship between pixels in the input image and modeling uncertainty of the prediction.
Those two main advantageous are brought by designing the target of our network to the lixel-based 1D heatmap.
This can make training much stable, and the system achieves much lower test error. 

\begin{figure}[t]
\begin{center}
\includegraphics[width=0.6\linewidth]{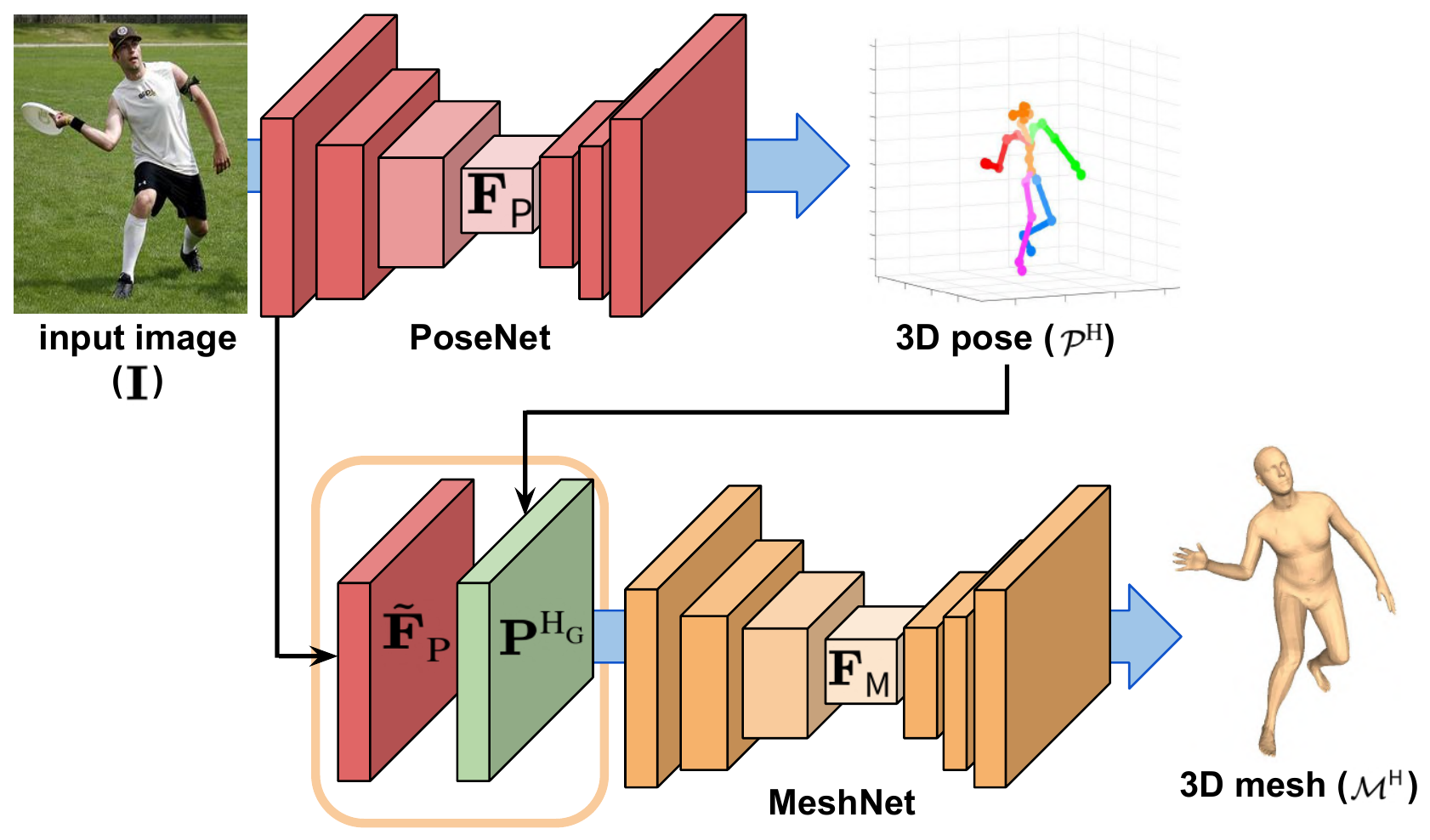}
\end{center}
   \caption{
   Overall pipeline of the proposed I2L-MeshNet. 
   }
\label{fig:overall_pipeline}
\end{figure}

\noindent \textbf{Heatmap-based 3D human pose estimation.}
Most of the recent state-of-the-art 2D and 3D human pose estimation methods use heatmap as a prediction target, which preserves the spatial relationship in the input image and models the uncertainty of the prediction. 
Tompson~et al.~\cite{tompson2014joint} proposed to estimate the Gaussian heatmap instead of directly regressing coordinates of human body joints. 
Their heatmap representation helps their model to perform 2D human pose estimation more accurate and motivated many heatmap-based 2D human pose methods~\cite{newell2016stacked,chen2018cascaded,xiao2018simple}. 
Pavlakos~et al.~\cite{pavlakos2017coarse} and Moon~et al.~\cite{moon2018v2v} firstly proposed to use 3D heatmaps as a prediction target for 3D human body pose and 3D hand pose estimation, respectively. 
Especially, Moon~et al.~\cite{moon2018v2v} demonstrated that under the same setting, changing prediction target from coordinates to heatmap significantly improves the 3D hand pose accuracy while requires much less amount of the learnable parameters.
Recently, Moon~et al.~\cite{moon2019camera} achieved significantly better 3D multi-person pose estimation accuracy using 3D heatmap compared with previous coordinate regression-based methods~\cite{rogez2017lcr}.

%% file: src/i2l-meshnet.tex
\begin{figure}[t]
\begin{center}
\includegraphics[width=1.0\linewidth]{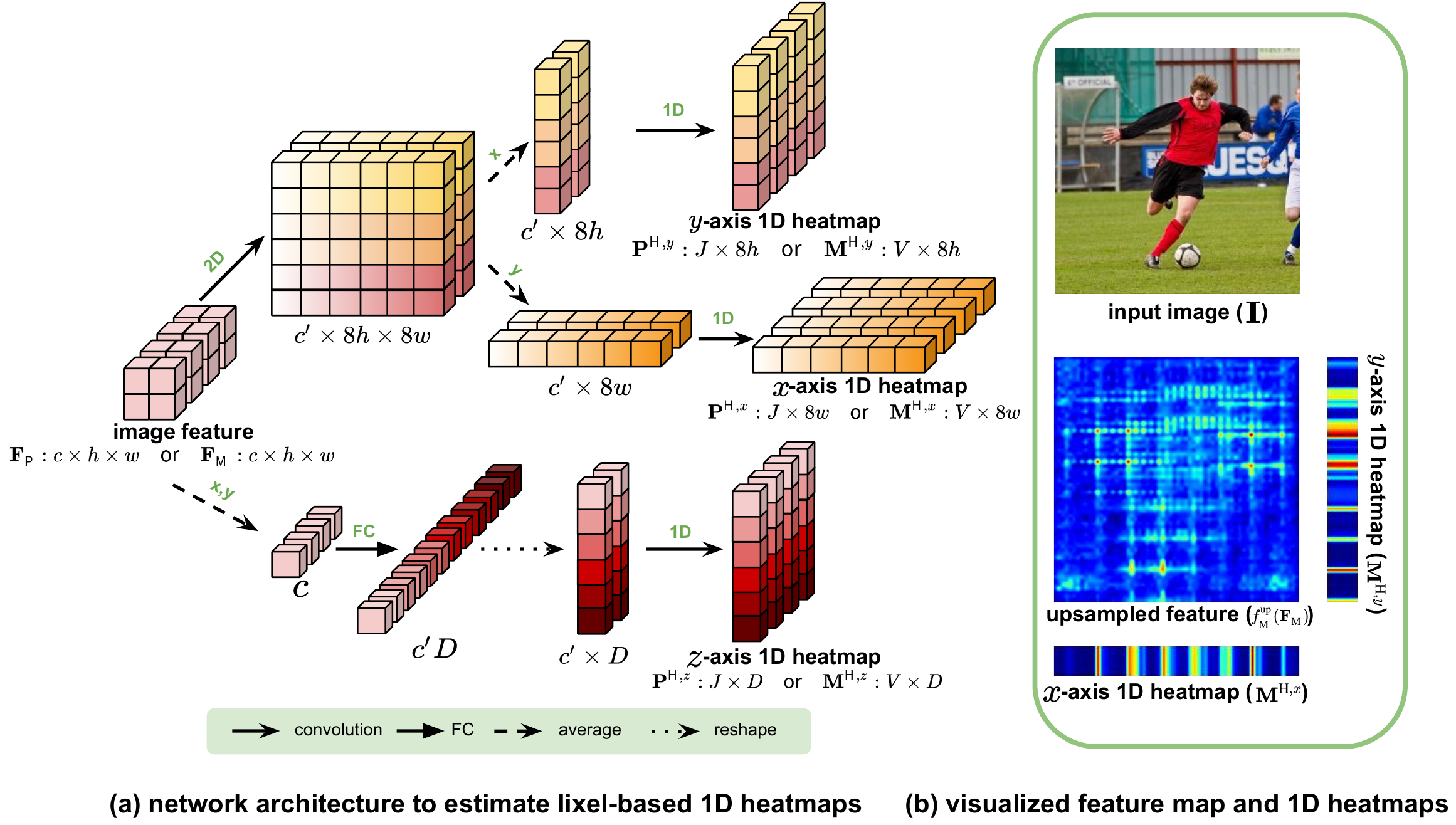}
\end{center}
   \caption{
   Network architecture to predict lixel-based 1D heatmaps and visualized examples of feature maps and the 1D heatmaps.
   }
\label{fig:lixel_1d_heatmap}
\end{figure}

\section{I2L-MeshNet}
Figure~\ref{fig:overall_pipeline} shows the overall pipeline of the proposed I2L-MeshNet.
I2L-MeshNet consists of PoseNet and MeshNet, which will be described in the following subsections.

\subsection{PoseNet}
The PoseNet estimates three lixel-based 1D heatmaps of all human joints $\mathcal{P}^\text{H} = \{\mathbf{P}^{\text H,x}, \mathbf{P}^{\text H,y}, \mathbf{P}^{\text H,z}\}$ from the input image $\textbf{I}$. 
$\mathbf{P}^{\text H,x}$ and $\mathbf{P}^{\text H,y}$ are defined in $x$- and $y$-axis of the image space, while $\mathbf{P}^{\text H,z}$ is defined in root joint (\textit{i.e.}, pelvis or wrist)-relative depth space.
For this, PoseNet extracts image feature $\mathbf{F}_\text P \in \mathbb{R}^{c \times h \times w}$ from the input image by ResNet~\cite{he2016deep}.
Then, three upsampling modules increases the spatial size of $\mathbf{F}_\text P$ by 8 times, while changing channel dimension from $c=2048$ to $c'=256$.
Each upsampling module consists of deconvolutional layer, 2D batch normalization layer~\cite{ioffe2015batch}, and ReLU function.
The upsampled features are used to compute lixel-based 1D human pose heatmaps, as illustrated in Figure~\ref{fig:lixel_1d_heatmap} (a).
We obtain $x$- and $y$-axis 1D human pose heatmaps as follows:
\begin{equation}
\mathbf{P}^{\text H,x} = f^{\text{1D},x}_\text P(\text{avg}^y(f^\text{up}_\text P(\mathbf{F}_\text P)))
\quad \text{and} \quad
\mathbf{P}^{\text H,y} = f^{\text{1D},y}_\text P(\text{avg}^x(f^\text{up}_\text P(\mathbf{F}_\text P))),
\end{equation}
where $f^\text{up}_\text P$ denotes the three upsampling modules of the PoseNet.
$\text{avg}^i$ and $f^{\text{1D},i}_\text P$ denote $i$-axis marginalization by averaging and a 1-by-1 1D convolution that changes channel dimension from $c'$ to $J$ for $i$-axis 1D human pose heatmap estimation, respectively.

We obtain $z$-axis 1D human pose heatmaps as follows:
\begin{equation}
\mathbf{P}^{\text H,z} = f^{\text{1D},z}_\text P(\psi(f_\text P(\text{avg}^{x,y}(\mathbf{F}_\text P)))),
\end{equation}
where $f_\text P$ and $\psi \colon \mathbb{R}^{c' D} \to \mathbb{R}^{c' \times D}$ denote a building block and and reshape function, respectively.
The building block consists of a fully-connected layer, 1D batch normalization layer, and ReLU function, and it changes the activation size from $c$ to $c'D$.
$D$ denotes depth discretization size and is equal to $8h=8w$.
We convert the discretized heatmaps of $\mathcal{P}^\text{H}$ to continuous coordinates $\mathbf{P}^\text C = [\mathbf{p}^{\text C, x}, \mathbf{p}^{\text C, y}, \mathbf{p}^{\text C, z}] \in \mathbb{R}^{J \times 3}$ by soft-argmax~\cite{sun2018integral}.

\subsection{MeshNet}
The MeshNet has a similar network architecture with that of the PoseNet.
Instead of taking the input image $\mathbf{I}$, MeshNet takes a pre-computed image feature from the PoseNet $\tilde{\mathbf{F}}_{\text P}$ and 3D Gaussian heatmap $\mathbf{P}^{\text H_\text G} \in \mathbb{R}^{J \times D \times 8h \times 8w}$.
$\tilde{\mathbf{F}}_{\text P}$ is the input of the first residual block of the PoseNet whose spatial dimension is $8h \times 8w$.
$\mathbf{P}^{\text H_\text G}$ is obtained from $\mathbf{P}^\text C$ as follows:
\begin{equation}
\mathbf{P}^{\text H_\text G} (j,z,y,x) =  \exp{\left(-\frac{(x-\mathbf{p}^{\text C,x}_j)^2+(y-\mathbf{p}^{\text C,y}_j)^2+(z-\mathbf{p}^{\text C,z}_j)^2}{2\sigma^2}\right)},
\end{equation}
where $\mathbf{p}^{\text C,x}_j$, $\mathbf{p}^{\text C,y}_j$ and $\mathbf{p}^{\text C,z}_j$ are $j$th joint $x$-, $y$-, and $z$-axis coordinates from $\mathbf{P}^\text C$, respectively.
$\sigma$ is set to 2.5.

From $\mathbf{P}^{\text H_\text G}$ and $\tilde{\mathbf{F}}_{\text P}$, we obtain image feature $\mathbf{F}_\text M$ as follows:
\begin{equation}
\mathbf{F}_\text M = \text{ResNet}_\text M (f_\text M(\psi(\mathbf{P}^{\text H_ \text G}) \oplus \tilde{\mathbf{F}}_{\text P})),
\end{equation}
where $\psi \colon \mathbb{R}^{J \times D \times 8h \times 8w} \to \mathbb{R}^{JD \times 8h \times 8w}$ and $\oplus$ denote reshape function and concatenation along the channel dimension, respectively.
$f_\text M$ is a convolutional block that consists of a 3-by-3 convolutional layer, 2D batch normalization layer, and ReLU function.
It changes the channel dimension of the input to the input channel dimension of the first residual block of the ResNet.
$\text{ResNet}_\text M$ is the ResNet starting from the first residual block.

From the $\mathbf{F}_\text M$, MeshNet outputs three lixel-based 1D heatmaps of all mesh vertices $\mathcal{M}^\text{H} = \{\mathbf{M}^{\text H,x}, \mathbf{M}^{\text H,y}, \mathbf{M}^{\text H,z}\}$ in an exactly the same manner with that of PoseNet, as illustrated in Figure~\ref{fig:lixel_1d_heatmap} (a).
Likewise heatmaps of PoseNet, $\mathbf{M}^{\text H,x}$ and $\mathbf{M}^{\text H,y}$ are defined in $x$- and $y$-axis of the image space, while $\mathbf{M}^{\text H,z}$ is defined in root joint-relative depth space.
We obtain $x$- and $y$-axis 1D human mesh heatmaps as follows:
\begin{equation}
\mathbf{M}^{\text H,x} = f^{\text{1D},x}_\text M(\text{avg}^y(f^\text{up}_\text M(\mathbf{F}_\text M)))
\quad \text{and} \quad
\mathbf{M}^{\text H,y} = f^{\text{1D},y}_\text M(\text{avg}^x(f^\text{up}_\text M(\mathbf{F}_\text M))),
\end{equation}
where $f^\text{up}_\text M$ denotes the three upsampling modules of the MeshNet.
$f^{\text{1D},i}_\text M$ denote a 1-by-1 1D convolution that changes channel dimension from $c'$ to $V$ for $i$-axis 1D human mesh heatmap estimation, respectively.
Figure~\ref{fig:lixel_1d_heatmap} (b) shows visualized $f_\text M^\text{up}(\mathbf{F}_\text M)$, $\mathbf{M}^\text{H,x}$, and $\mathbf{M}^\text{H,y}$.

We obtain $z$-axis 1D human mesh heatmaps as follows:
\begin{equation}
\mathbf{M}^{\text H,z} = f^{\text{1D},z}_\text M(\psi(f_\text M(\text{avg}^{x,y}(\mathbf{F}_\text M)))),
\end{equation}
where $f_\text M$ and $\psi \colon \mathbb{R}^{c' D} \to \mathbb{R}^{c' \times D}$ denote a building block and and reshape function, respectively.
The building block consists of a fully-connected layer, 1D batch normalization layer, and ReLU function, and it changes the activation size from $c$ to $c'D$.
Likewise we did in the PoseNet, we convert the discretized heatmaps of $\mathcal{M}^\text{H}$ to continuous coordinates $\mathbf{M}^\text C = [\mathbf{m}^{\text C, x}, \mathbf{m}^{\text C, y}, \mathbf{m}^{\text C, z}] \in \mathbb{R}^{V \times 3}$ by soft-argmax~\cite{sun2018integral}.

\subsection{Final 3D human pose and mesh}
The final 3D human mesh $\mathbf{M}$ and pose $\mathbf{P}$ are obtained as follows:
\begin{equation}
\mathbf{M} = \mathrm{\Pi}(\mathbf{T}^{-1}\mathbf{M}^{\text C} + \mathbf{R})
\quad \text{and} \quad
\mathbf{P} = \mathcal{J}\mathbf{M},
\end{equation}
\noindent where $\mathrm{\Pi}$, $\mathbf{T}^{-1}$, and $\mathbf{R} \in \mathbb{R}^{1 \times 3}$ denote camera back-projection, inverse affine transformation (\textit{i.e.}, 2D crop and resize), and $z$-axis offset whose element is a depth of the root joint, respectively.
$\mathbf{R}$ is obtained from RootNet~\cite{moon2019camera}.
We use normalized camera intrinsic parameters if not available following Moon~et al.~\cite{moon2019camera}.
$\mathcal{J} \in \mathbb{R}^{J \times V}$ is a joint regression matrix defined in SMPL or MANO model.

\subsection{Loss functions} ~\label{section:loss_functions}

\textbf{PoseNet pose loss.}
To train the PoseNet, we use $L1$ loss function defined as follows:
\begin{equation}
L_\text{pose}^\text{PoseNet} = \| \mathbf{P}^\text C - \mathbf{P}^{\text C*} \|_1 ,  
\end{equation}
where $*$ indicates groundtruth. 
$z$-axis loss becomes zero if $z$-axis groundtruth is unavailable.

\textbf{MeshNet pose loss.}
To train the MeshNet to predict mesh vertex aligned with body joint locations, we use $L1$ loss function defined as follows:
\begin{equation}
L_\text{pose}^\text{MeshNet} = \| \mathcal{J}\mathbf{M}^\text C - \mathbf{P}^{\text C*} \|_1 ,  
\end{equation}
where $*$ indicates groundtruth. 
$z$-axis loss becomes zero if $z$-axis groundtruth is unavailable.

\textbf{Mesh vertex loss.}
To train the MeshNet to output mesh vertex heatmaps, we use $L1$ loss function defined as follows:
\begin{equation}
L_\text {vertex} = \| \mathbf{M}^\text C - \mathbf{M}^{\text C*} \|_1 ,  
\end{equation}
where $*$ indicates groundtruth. 
$z$-axis loss becomes zero if $z$-axis groundtruth is unavailable.

\textbf{Mesh normal vector loss.}
Following Wang~et al.~\cite{wang2018pixel2mesh}, we supervise normal vector of predicted mesh to get visually pleasing mesh result. The $L1$ loss function for normal vector supervision is defined as follows:
\begin{equation}
L_\text{normal} = \sum_f \sum_{\{i,j\} \subset f} 
                \Big |  \Big \langle
                    \frac{\mathbf{m}_i^\text C - \mathbf{m}_j^\text C}
                         {\| \mathbf{m}_i^\text C - \mathbf{m}_j^\text C \|_2 } , 
                    n_f^* 
                    \Big \rangle
                \Big | ,  
\end{equation}
where $f$ and $n_f$ indicate a mesh face and unit normal vector of face $f$, respectively. 
$\mathbf{m}_i^\text C$ and $\mathbf{m}_j^\text C$ denote $i$th and $j$th vertex coordinates of $\mathbf{M}^\text C$, respectively.
$n_f^*$ is computed from $\mathbf{M}^{\text C*}$, where $*$ denotes groundtruth. 
The loss becomes zero if groundtruth 3D mesh is unavailable.

\textbf{Mesh edge length loss.}
Following Wang~et al.~\cite{wang2018pixel2mesh}, we supervise edge length of predicted mesh to get visually pleasing mesh result. The $L1$ loss function for edge length supervision is defined as follows:
\begin{equation}
L_\text{edge} = \sum_f \sum_{\{i,j\} \subset f} = | 
        \|\mathbf{m}_i^\text C - \mathbf{m}_j^\text C \|_2 
        - \|\mathbf{m}_i^{\text C*} - \mathbf{m}_j^{\text C*} \|_2 
                                        | ,  
\end{equation}
where $f$ and $*$ indicate mesh face and groundtruth, respectively. 
$\mathbf{m}_i^\text C$ and $\mathbf{m}_j^\text C$ denote $i$th and $j$th vertex coordinates of $\mathbf{M}^\text C$, respectively.
The loss becomes zero if groundtruth 3D mesh is unavailable.

We train our I2L-MeshNet in an end-to-end manner using all the five loss functions as follows:
\begin{equation}
L = L_\text{pose}^\text{PoseNet} + L_\text{pose}^\text{MeshNet} + L_\text{vertex} + \lambda L_\text{normal} + L_\text{edge},  
\end{equation}
where $\lambda=0.1$ is a weight of $L_\text{normal}$.
For the stable training, we do not back-propagate gradients before $\mathbf{P}^{\text H_\text G}$.

%% file: src/implementation_details.tex
\section{Implementation details}~\label{section:implementation_details}

PyTorch~\cite{paszke2017automatic} is used for implementation. 
The backbone part is initialized with the publicly released ResNet-50~\cite{he2016deep} pre-trained on the ImageNet dataset~\cite{russakovsky2015imagenet}, and the weights of the remaining part are initialized by Gaussian distribution with $\sigma=0.001$. 
The weights are updated by the Adam optimizer~\cite{kingma2014adam} with a mini-batch size of 48. 
To crop the human region from the input image, we use groundtruth bounding box in both of training and testing stages following previous works~\cite{kanazawa2018end,kolotouros2019convolutional,kolotouros2019learning}.
When the bounding box is not available in the testing stage, we trained and tested Mask R-CNN~\cite{he2017mask} to get the bounding box.
The cropped human image is resized to 256$\times$256, thus $D=64$ and $h=w=8$.
Data augmentations including scaling ($\pm$25\%), rotation ($\pm$\ang{60}), random horizontal flip, and color jittering ($\pm$20\%) is performed in training.
The initial learning rate is set to $10^{-4}$ and reduced by a factor of 10 at the \nth{10} epoch.
We train our model for 12 epochs with three NVIDIA RTX 2080Ti GPUs, which takes 36 hours for training. 
Our I2L-MeshNet runs at a speed of 25 frames per second (fps).

%% file: src/experiment.tex
\section{Experiment}

\subsection{Datasets and evaluation metrics} ~\label{section:dataset}

\noindent \textbf{Human3.6M.}
Human3.6M~\cite{ionescu2014human3} contains 3.6M video frames with 3D joint coordinate annotations. 
Because of the license problem, previously used groundtruth SMPL parameters of the Human3.6M are inaccessible.
Alternatively, we used SMPLify-X~\cite{pavlakos2019expressive} to obtain groundtruth SMPL parameters.
Please see the supplementary material for a detailed description of SMPL parameters of the Human3.6M.
MPJPE and PA MPJPE are used for the evaluation~\cite{moon2019camera}, which is Euclidean distance (mm) between predicted and groundtruth 3D joint coordinates after root joint alignment and further rigid alignment, respectively.

\noindent \textbf{3DPW.}
3DPW~\cite{von2018recovering} contains 60 video sequences captured mostly in outdoor conditions. 
We use this dataset only for evaluation on its defined test set following Kolotouros~et al.~\cite{kolotouros2019learning}.
The same evaluation metrics with Human3.6M (\textit{i.e.}, MPJPE and PA MPJPE) are used, following Kolotouros~et al.~\cite{kolotouros2019learning}.

\noindent \textbf{FreiHAND.}
FreiHAND~\cite{Freihand2019} contains real-captured 130K training images and 4K test images with MANO pose and shape parameters.
The evaluation is performed at an online server. 
Following Zimmermann~et al.~\cite{Freihand2019}, we report PA MPVPE, PA MPJPE, and F-scores.

\noindent \textbf{MSCOCO.}
MSCOCO~\cite{lin2014microsoft} contains large-scale in-the-wild images with 2D bounding box and human joint coordinates annotations.
We fit SMPL using SMPLify-X~\cite{pavlakos2019expressive} on the groundtruth 2D poses, and used the fitted meshes as groundtruth 3D meshes.
This dataset is used only for the training.

\noindent \textbf{MuCo-3DHP.}
MuCo-3DHP~\cite{mehta2018single} is generated by compositing the existing MPI-INF-3DHP 3D~\cite{mehta2017monocular}.
200K frames are composited, and half of them have augmented backgrounds.
We used images of MSCOCO dataset that do not include humans to augment the backgrounds following Moon~et al.~\cite{moon2019camera}.
This dataset is used only for the training.

\subsection{Ablation study}
All models for the ablation study are trained and tested on Human3.6M. 
As Human3.6M is the most widely used large-scale benchmark, we believe this dataset is suitable for the ablation study. 

\noindent \textbf{Benefit of the heatmap-based mesh estimation.}
To demonstrate the benefit of the heatmap-based mesh estimation, we compare models with various target representations of the human mesh, such as SMPL parameters, vertex coordinates, and heatmap. 
Table~\ref{table:ablation_hm} shows MPJPE, the number of parameters, and the GPU memory usage comparison between models with different targets.
The table shows that our heatmap-based mesh estimation network achieves the lowest errors while using the smallest number of the parameters and consuming small GPU memory.

The superiority of our heatmap-based mesh estimation network is in two folds.
First, it can model the uncertainty of the prediction.
To validate this, we trained two models that estimate the camera-centered mesh vertex coordinates directly and estimates lixel-based 1D heatmap of the coordinates using two fully-connected layers.
Note that the targets of the two models are the same, but their representations are different.
As the first network regresses the coordinates directly, it cannot model the uncertainty on the prediction, while the latter one can because of the heatmap target representation.
However, both do not preserve the spatial relationship in the input image because of the global average pooling and the fully-connected layers.
As the second and third row of the table show, modeling uncertainty on the prediction significantly decreases the errors while using a smaller number of parameters.
In addition, it achieves lower errors than the SMPL parameter regression model, which is the most widely used target representation but cannot model the uncertainty.

\begin{table}[t]
\centering
\setlength\tabcolsep{1.0pt}
\def\arraystretch{1.1}
\scalebox{1.0}{
\begin{tabular}{C{4.0cm}|C{1.0cm}C{2.0cm}C{1.5cm}C{1.8cm}C{1.6cm}}
\specialrule{.1em}{.05em}{.05em}
targets & spatial & uncertainty & MPJPE & no. param. & GPU mem.\\ \hline
SMPL param. & \xmark & \xmark & 100.3 & 91M & 4.3 GB \\
xyz coord. & \xmark & \xmark & 114.3 & 117M & 5.4 GB\\
xyz lixel hm. wo. spatial & \xmark & \cmark & 92.6 & 82M & 4.5 GB\\
\textbf{xyz lixel hm. (ours)} & \cmark & \cmark & \textbf{86.2} & \textbf{73M} & \textbf{4.6 GB} \\
 \specialrule{.1em}{.05em}{.05em}
\end{tabular}
}
\caption{The MPJPE, the number of parameters, and the GPU memory usage comparison between various target representations on Human3.6M.}
\label{table:ablation_hm}
\end{table}

\begin{table}[t]
\centering
\setlength\tabcolsep{1.0pt}
\def\arraystretch{1.1}
\scalebox{1.0}{
\begin{tabular}{C{4.3cm}|C{2.2cm}C{1.8cm}C{1.3cm}C{1.8cm}}
\specialrule{.1em}{.05em}{.05em}
targets & mem. complx. & resolution & MPJPE & GPU mem.\\ \hline
\multirow{2}{*}{xyz voxel hm.} & \multirow{2}{*}{$\mathcal{O}(VD^3)$} & 8$\times$8$\times$8 & 102.8 & 4.3 GB\\
& & 16$\times$16$\times$16 & - & OOM \\ \hline
\multirow{3}{*}{xy pixel hm. + z lixel hm.} & \multirow{3}{*}{$\mathcal{O}(VD^2)$} & 8$\times$8, 8 & 97.9 & 3.5 GB\\
& & 32$\times$32, 32 & 89.4 & 5.7 GB\\
& & 64$\times$64, 64 & - & OOM \\ \hline
\multirow{3}{*}{\textbf{xyz lixel hm. (ours)}} & \multirow{3}{*}{$\mathcal{O}(VD)$} & 8, 8, 8 & 100.2 & 3.4 GB\\
& & 32, 32 ,32 & 94.8 & 4.0 GB\\
& & 64, 64, 64 & \textbf{86.2} & \textbf{4.6 GB} \\
 \specialrule{.1em}{.05em}{.05em}
\end{tabular}
}
\caption{The MPJPE and the GPU memory usage comparison between various heatmap representations on Human3.6M.}
\label{table:ablation_lixel}
\end{table}

Second, it preserves the spatial relationship between pixels in the input image.
The final model estimates the $x$- and $y$-axis heatmaps of each mesh vertex in a fully-convolutional way, thus preserves the spatial relationship.
It achieves the best performance with the smallest number of the parameters while consuming similar GPU memory usage compared with SMPL parameter regression method that requires the least amount of GPU memory.

In Table~\ref{table:ablation_hm}, all models have the same network architecture with our I2L-MeshNet except for the final output prediction part.
We removed PoseNet from all models, and the remaining MeshNet directly estimates targets from the input image $\mathbf{I}$.
Except for the last row (ours), all settings output targets using two fully-connected layers.
We followed the training details of ~\cite{kanazawa2018end,kolotouros2019learning} for the SMPL parameter estimation.

\begin{table}[t]
\centering
\setlength\tabcolsep{1.0pt}
\def\arraystretch{1.1}
\scalebox{1.0}{
\begin{tabular}{C{4.6cm}|C{1.3cm}C{1.5cm}C{1.7cm}}
\specialrule{.1em}{.05em}{.05em}
settings & 3D pose & MPJPE & PA MPJPE  \\ \hline
MeshNet & \xmark & 86.2 & 59.8 \\
\textbf{PoseNet+MeshNet (ours)} & \cmark & \textbf{81.8} & \textbf{58.0} \\ \hline
MeshNet & GT & 25.5 & 17.1 \\
 \specialrule{.1em}{.05em}{.05em}
\end{tabular}
}
\caption{The MPJPE and PA MPJPE comparison between various network cascading strategies on Human3.6M.}
\label{table:ablation_3dpose}
\end{table}

\noindent \textbf{Lixel-based vs. pixel-based vs. voxel-based heatmap.}
To demonstrate the effectiveness of the lixel-based 1D heatmap over other heatmap representations, we train three models that predict lixel-based, pixel-based, and voxel-based heatmap, respectively.
We used the same network architecture (\textit{i.e.}, MeshNet of the I2L-MeshNet) for all settings except for the final prediction part.
Their networks directly predict the heatmaps from the input image.
$x$-, $y$-, and $z$-axis of each heatmap represents the same coordinates.
Table~\ref{table:ablation_lixel} shows memory complexity, heatmap resolution, MPJPE and GPU memory usage comparison between models that predict different target representations of human mesh. 
The table shows that our lixel-based one achieves the lowest error while consuming small GPU memory usage.

Compared with the pixel-based and voxel-based heatmap, our lixel-based one consumes much less amount of GPU memory under the same resolution.
The $8 \times 8 \times 8$ voxel-based heatmap requires similar GPU memory usage with that of $64, 64, 64$ lixel-based one, and we found that enlarging the voxel-based heatmap size from it is not allowed in current GPU memory limit (\textit{i.e.}, 12 GB).
The pixel-based heatmap is more efficient than the voxel-based one; however still much inefficient than our lixel-based one, which makes enlarging from $32 \times 32, 32$ impossible.
This inefficient memory usage limits the heatmap resolution; however, we found that the heatmap resolution is critical for dense mesh vertex localization.
On the other hand, the memory complexity of our lixel-based heatmap is a linear function with respect to $D$; thus, we can predict high-resolution heatmap for each mesh vertex.
The memory efficiency will be more important when a high-resolution human mesh model is used.

Under the same resolution, the combination of pixel-based heatmap and lixel-based heatmap achieves the best performance.
We think that estimating the voxel-based heatmap involves too many parameters at a single output layer, which makes it produce high errors.
In addition, lixel-based heatmap inherently involves spatial ambiguity arises from marginalizing the 2D feature map to 1D, which can be a possible reason for worse performance than the combined one.

\begin{table}[t]
\setlength{\tabcolsep}{1pt}
\centering
\scalebox{1.0}{
\begin{tabular}{C{4.0cm}|C{1.5cm}C{1.7cm}|C{1.5cm}C{1.7cm}}
\specialrule{.1em}{.05em}{.05em}
\multirow{ 2}{*}{methods} & \multicolumn{2}{c|}{Human3.6M} & \multicolumn{2}{c}{3DPW} \\
                       & MPJPE & PA MPJPE & MPJPE & PA MPJPE\\ \hline
HMR~\cite{kanazawa2018end} & 153.2 & 85.5 & 300.4 & 137.2 \\
GraphCMR~\cite{kolotouros2019convolutional} & 78.3 & 59.5 & 126.5 & 80.1 \\
SPIN~\cite{kolotouros2019learning} & 72.9 & 51.9 & 113.1 & 71.7 \\
\textbf{I2L-MeshNet (Ours)} & \textbf{55.7} & \textbf{41.7} & \textbf{95.4} & \textbf{60.8} \\
 \specialrule{.1em}{.05em}{.05em}
\end{tabular}
}
\caption{The MPJPE and PA MPJPE comparison on Human3.6M and 3DPW. All methods are trained on Human3.6M and MSCOCO.}
\label{table:compare_h36m_3dpw_same_dataset}
\end{table}

\begin{table}[t]
\setlength{\tabcolsep}{1pt}
\begin{minipage}{.45\linewidth}
\centering

\scalebox{1.0}{
\begin{tabular}{C{2.5cm}|C{1.3cm}C{1.7cm}}
\specialrule{.1em}{.05em}{.05em}
methods &  MPJPE & PA MPJPE \\ \hline
SMPLify~\cite{bogo2016keep} & - & 82.3 \\
Lassner~\cite{lassner2017unite} &- & 93.9 \\
HMR~\cite{kanazawa2018end} & 88.0 & 56.8  \\
NBF~\cite{omran2018neural} & - & 59.9 \\
Pavlakos~\cite{pavlakos2018learning} & - & 75.9  \\
Kanazawa~\cite{kanazawa2019learning} & - & 56.9  \\
GraphCMR~\cite{kolotouros2019convolutional} & - & 50.1  \\
Arnab~\cite{arnab2019exploiting} &77.8 & 54.3  \\
SPIN~\cite{kolotouros2019learning} & - & \textbf{41.1}  \\
\textbf{I2L-MeshNet (Ours)} & \textbf{55.7} & \textbf{41.1} \\
 \specialrule{.1em}{.05em}{.05em}
\end{tabular}
}
\caption{The MPJPE and PA MPJPE comparison on Human3.6M. Each method is trained on different datasets.}
\label{table:compare_h36m_different_dataset}
\end{minipage}\hfill
\begin{minipage}{.50\linewidth}
\centering
\scalebox{1.0}{
\begin{tabular}{C{2.5cm}|C{1.3cm}C{1.7cm}}
\specialrule{.1em}{.05em}{.05em}
methods & MPJPE & PA MPJPE \\ \hline
HMR~\cite{kanazawa2018end} & - & 81.3 \\
Kanazawa~\cite{kanazawa2019learning}  & - & 72.6 \\
GraphCMR~\cite{kolotouros2019convolutional} & - & 70.2 \\
Arnab~\cite{arnab2019exploiting}  & - & 72.2 \\
SPIN~\cite{kolotouros2019learning}  & - & 59.2 \\
\textbf{I2L-MeshNet (Ours)} & \textbf{93.2} & \textbf{57.7} \\
\textbf{I2L-MeshNet (Ours) + SMPL regress} & 100.0 & 60.0 \\
 \specialrule{.1em}{.05em}{.05em}
\end{tabular}
}
\caption{The MPJPE and PA MPJPE comparison on 3DPW. Each method is trained on different datasets.}
\label{table:compare_3dpw_different_dataset}
\end{minipage}
\end{table}

\noindent \textbf{Benefit of the cascaded PoseNet and MeshNet.}
To demonstrate the benefit of the cascaded PoseNet and MeshNet, we trained and tested three networks using various network cascading strategy.
First, we removed PoseNet from the I2L-MeshNet. 
The remaining MeshNet directly predicts lixel-based 1D heatmap of each mesh vertex from the input image.
Second, we trained I2L-MeshNet, which has cascaded PoseNet and MeshNet architecture.
Third, to check the upper bound accuracy with respect to the output of the PoseNet, we fed the groundtruth 3D human pose instead of the output of the PoseNet to the MeshNet in both training and testing stage.
Table~\ref{table:ablation_3dpose} shows utilizing the output of the PoseNet (the second row) achieves better accuracy compared with using only MeshNet (the first row) to estimate the human mesh.
Interestingly, passing the groundtruth 3D human pose to the MeshNet (the last row) significantly improves the performance compared with all the other settings.
This indicates that improving the 3D human pose estimation network can be one important way to improve 3D human mesh estimation accuracy.

\subsection{Comparison with state-of-the-art methods}

\begin{table}[t]
\centering
\setlength\tabcolsep{1.0pt}
\def\arraystretch{1.1}
\scalebox{1.0}{
\begin{tabular}{C{3.5cm}|C{1.8cm}C{1.8cm}C{1.5cm}C{1.5cm}C{1.5cm}}
\specialrule{.1em}{.05em}{.05em}
 methods &  PA MPVPE &  PA MPJPE &  F@5 mm &  F@15 mm & GT scale \\ \hline
 Hasson~et al.~\cite{hasson2019learning} & 13.2 & - & 0.436 & 0.908 & \cmark \\
Boukhayma~et al.~\cite{boukhayma20193d} & 13.0 & - & 0.435 & 0.898 & \cmark \\
FreiHAND~\cite{Freihand2019} & 10.7 & - & \scriptsize0.529 & 0.935 & \cmark \\
\textbf{I2L-MeshNet (Ours)} & \textbf{7.6} & \textbf{7.4} & \textbf{0.681} & \textbf{0.973} & \xmark \\
 \specialrule{.1em}{.05em}{.05em}
\end{tabular}
}
\caption{
The PA MPVPE, PA MPJPE, and F-scores comparison between state-of-the-art methods and the proposed I2L-MeshNet on FreiHAND.
The checkmark denotes a method use groundtruth information during inference time.
}
\label{table:compare_freihand}
\end{table}

\noindent \textbf{Human3.6M and 3DPW.}
We compare the MPJPE and PA MPJPE of our I2L-MeshNet with previous state-of-the-art 3D human body pose and mesh estimation methods on Human3.6M and 3DPW test set.
As each previous work trained their network on different training sets, we report the 3D errors in two ways.

First, we train all methods on Human3.6M and MSCOCO and report the errors in Table~\ref{table:compare_h36m_3dpw_same_dataset}.
The previous state-of-the-art methods~\cite{kanazawa2018end,kolotouros2019convolutional,kolotouros2019learning} are trained from their officially released codes.
The table shows that our I2L-MeshNet significantly outperforms previous methods by a large margin on both datasets.

Second, we report the 3D errors of previous methods from their papers and ours in Table~\ref{table:compare_h36m_different_dataset} and Table~\ref{table:compare_3dpw_different_dataset}. 
Each network of the previous method is trained on the different combinations of datasets, which include Human3.6M, MSCOCO, MPII~\cite{andriluka20142d}, LSP~\cite{johnson2010clustered}, LSP-Extended~\cite{johnson2011learning}, UP~\cite{lassner2017unite}, and MPI-INF-3DHP~\cite{mehta2017monocular}.
We used MuCo-3DHP for the additional training dataset for the evaluation on 3DPW dataset.
We also report the 3D errors from a additional SMPL parameter regression module following Kolotouros~et al.~\cite{kolotouros2019convolutional}.
The tables show that the performance gap between ours and the previous state-of-the-art method~\cite{kolotouros2019learning} is significantly reduced.

The reason for the reduced performance gap is that previous model-based state-of-the-art methods~\cite{kanazawa2018end,kolotouros2019learning} can get benefit from many in-the-wild 2D human pose datasets~\cite{lin2014microsoft,johnson2010clustered,johnson2011learning} by a 2D pose-based weak supervision.
As the human body or hand model assumes a prior distribution between the human model parameters (\textit{i.e.}, 3D joint rotations and identity vector) and 3D joint/mesh coordinates, the 2D pose-based weak supervision can provide gradients in depth axis, calculated from the prior distribution.
Although the weak supervision still suffers from the depth ambiguity, utilizing in-the-wild images can be highly beneficial because the images have diverse appearances compared with those of the lab-recorded 3D datasets~\cite{ionescu2014human3,mehta2017monocular,mehta2018single}.
On the other hand, model-free approaches, including the proposed I2L-MeshNet, do not assume any prior distribution, therefore hard to get benefit from the weak supervision.
Based on the two comparisons, we can draw two important conclusions.
\begin{itemize}
\item I2L-MeshNet achieve much higher accuracy than the model-based methods when trained on the same datasets that provide groundtruth 3D human poses and meshes.
\item The model-based approaches can achieve comparable or higher accuracy by utilizing additional in-the-wild 2D pose data without requiring the 3D supervisions.
\end{itemize}

We think that a larger number of accurately aligned in-the-wild image-3D mesh data can significantly boost the accuracy of I2L-MeshNet.
The iterative fitting~\cite{bogo2016keep,pavlakos2019expressive}, neural network~\cite{joo2020exemplar}, or their combination~\cite{kolotouros2019learning} can be used to obtain more data.
This can be an important future research direction, and we leave this as future work.

\noindent \textbf{FreiHAND.}
We compare MPVPE and F-scores of our I2L-MeshNet with previous state-of-the-art 3D human hand pose and mesh estimation methods~\cite{boukhayma20193d,hasson2019learning,Freihand2019}.
We trained Mask R-CNN~\cite{he2017mask} on FreiHAND train images to get the hand bounding box of test images.
Table~\ref{table:compare_freihand} shows that the proposed I2L-MeshNet significantly outperforms all previous works without groundtruth scale information during the inference time.
We additionally report MPJPE in the table.

%% file: src/conclusion.tex
\section{Conclusion}

We propose a I2L-MeshNet, image-to-lixel prediction network for accurate 3D human pose and mesh estimation from a single RGB image.
We convert the output of the network to the lixel-based 1D heatmap, which preserves the spatial relationship in the input image and models uncertainty of the prediction.
Our lixel-based 1D heatmap requires much less GPU memory usage under the same heatmap resolution while producing better accuracy compared with a widely used voxel-based 3D heatmap.
Our I2L-MeshNet outperforms previous 3D human pose and mesh estimation methods on various 3D human pose and mesh datasets.
We hope our method can give useful insight to the following model-free 3D human pose and mesh estimation approaches.

%% file: src/suppl.tex
\clearpage

\begin{center}
\textbf{\large Supplementary Material of \enquote{I2L-MeshNet: Image-to-Lixel Prediction Network for \\ Accurate 3D Human Pose and Mesh Estimation \\ from a Single RGB Image}}
\end{center}

In this supplementary material, we present more experimental results that could not be included in the main manuscript due to the lack of space.

\section{Various settings of I2L-MeshNet}
\subsection{When to marginalize 2D to 1D?}
We report how the MPJPE, PA MPJPE, and GPU memory usage change when the marginalization takes place on the ResNet output (\textit{i.e.}, $\mathbf{F}_\text P$ or $\mathbf{F}_\text M$), which is the input of the first upsampling module, instead of the output of the last upsampling module (\textit{i.e.}, $f^\text P_\text{up}(\textbf{F}_\text P)$ or $f^\text M_\text{up}(\textbf{F}_\text M)$) in Table~\ref{table:when_margi}.
For the convenience, we removed PoseNet from our I2L-MeshNet and changed MeshNet to take the input image.
The table shows that the early marginalization increases the errors while requiring less amount of GPU memory.
This is because the marginalized two 1D feature maps can be generated from multiple 2D feature map, which results in spatial ambiguity.
To reduce the effect of this spatial ambiguity, we designed our I2L-MeshNet to extract a sufficient amount of 2D information and then apply the marginalization at the last part of the network instead of applying it in the early stage.

When the marginalization is applied on the ResNet output $\mathbf{F}_\text M$, all 2D layers (\textit{i.e.}, deconvolutional layers and batch normalization layers) in the upsampling modules are converted to the 1D layers.
All models are trained on Human3.6M dataset.
The $z$-axis heatmap prediction part is not changed.

\begin{table}
\centering
\setlength\tabcolsep{1.0pt}
\def\arraystretch{1.1}
\scalebox{1.0}{
\begin{tabular}{C{4.0cm}|C{2.0cm}C{2.0cm}C{2.0cm}}
\specialrule{.1em}{.05em}{.05em}
settings & MPJPE & PA MPJPE & GPU mem.\\ \hline
avg on $\mathbf{F}_\text M$ & 93.5 & 64.1 & \textbf{4.4 GB} \\
\textbf{avg on $f^\text M_\text{up}(\textbf{F}_\text M)$ (ours)} & \textbf{86.2} & \textbf{59.8} & 4.6 GB\\ \hline
 \specialrule{.1em}{.05em}{.05em}
\end{tabular}
}
\caption{The MPJPE, PA MPJPE, and GPU memory usage comparison between various marginalization settings on Human3.6M dataset.}
\vspace*{-7mm}
\label{table:when_margi}
\end{table}

\subsection{How to marginalize 2D to 1D?}
We report how the MPJPE and PA MPJPE change when different marginalization methods are used in Table~\ref{table:how_margi}.
For the convenience, we removed PoseNet from our I2L-MeshNet and changed MeshNet to take the input image.
The table shows that our average pooling achieves the lowest errors.
Compared with the max pooling that provides the gradients to one pixel position per one $x$ or $y$ position, our average pooling provides the gradients to all pixel positions, which is much richer ones.
We implemented the weighted sum by constructing a convolutional layer whose kernel size is $(8h,1)$ and $(1,8w)$ for $x$- and $y$-axis lixel-based 1D heatmap prediction, respectively, without padding.
The weighted sum provides lower error than that of the max pooling, however still worse than our average pooling.
We believe the large size of a kernel of the convolutional layer (\textit{i.e.}, $(8h,1)$ and $(1,8w)$) is hard to be optimized, which results in higher error than ours.
For all settings, models are trained on Human3.6M dataset, and the $z$-axis heatmap prediction part is not changed.

\begin{table}
\centering
\setlength\tabcolsep{1.0pt}
\def\arraystretch{1.1}
\scalebox{1.0}{
\begin{tabular}{C{4.0cm}|C{2.0cm}C{2.0cm}}
\specialrule{.1em}{.05em}{.05em}
settings & MPJPE & PA MPJPE \\ \hline
max pooling & 93.5 & 64.1  \\
weighted sum & 89.4 & 61.4 \\
\textbf{avg pooling (ours)} & \textbf{86.2} & \textbf{59.8} \\ \hline
 \specialrule{.1em}{.05em}{.05em}
\end{tabular}
}
\caption{The MPJPE and PA MPJPE comparison between various marginalization settings on Human3.6M dataset.}
\vspace*{-7mm}
\label{table:how_margi}
\end{table}

\section{Comparison with previous 2.5D heatmap regression}
We compare the MPJPE and GPU memory usage between a model that predicts our lixel-based 1D heatmap and a model that predicts the 2.5D heatmap~\cite{iqbal2018hand} in Table~\ref{table:iqbal_lixel}.
The 2.5D heatmap~\cite{iqbal2018hand} consists of $xy$ heatmap and $z$ heatmap, where $xy$ one is the pixel-based 2D heatmap and $z$ one has the same spatial size with that of $xy$ heatmap and contains root joint-relative depth on the activated $xy$ position for all mesh vertices.
They predict the depth values on $z$ heatmap, not the likelihood, thus cannot model uncertainty of the $z$-axis prediction.
As the table shows, our lixel-based one achieves significantly lower error under the same resolution while requiring a much smaller amount of GPU memory.
We think that this is because the 2.5D heatmap of Iqbal~et al.~\cite{iqbal2018hand} cannot model uncertainty of the prediction in $z$-axis, while ours can.
For all settings, models are trained on Human3.6M dataset, and we removed PoseNet and changed MeshNet to take an input image and predict the heatmap.

\begin{table}
\centering
\setlength\tabcolsep{1.0pt}
\def\arraystretch{1.1}
\scalebox{0.8}{
\begin{tabular}{C{3.5cm}C{2.5cm}C{3.0cm}|C{1.4cm}C{1.8cm}}
\specialrule{.1em}{.05em}{.05em}
settings & resolution & uncertainty in $z$-axis & MPJPE & GPU mem. \\ \hline
2.5D heatmap~\cite{iqbal2018hand} & 8 $\times$ 8, 8 $\times$ 8 & \xmark & 107.4 & 3.6GB  \\
2.5D heatmap~\cite{iqbal2018hand} & 32 $\times$ 32, 32 $\times$ 32 & \xmark & 100.4 & 8.4GB  \\
lixel-based 1D heatmap & 8, 8, 8 & \cmark &  100.2 & 3.4GB  \\
lixel-based 1D heatmap & 32, 32, 32 & \cmark & 94.8 & 4.0GB  \\
lixel-based 1D heatmap & 64, 64, 64 & \cmark &  86.2 & 4.6GB  \\ \hline
 \specialrule{.1em}{.05em}{.05em}
\end{tabular}
}
\caption{The MPJPE and GPU memory usage comparison between various marginalization settings on Human3.6M dataset.}
\vspace*{-7mm}
\label{table:iqbal_lixel}
\end{table}

\section{Effect of each loss function}
We show the effectiveness of the MeshNet pose loss $L_\text{pose}^\text{MeshNet}$ in Table~\ref{table:meshnet_pose_loss}.
Although we supervise mesh vertices by the mesh vertex loss $L_\text{pose}^\text{MeshNet}$, additional $L_\text{pose}^\text{MeshNet}$ is helpful for human joint-aligned mesh prediction.
Both models are trained on Human3.6M dataset.

\begin{figure}
\begin{center}
\includegraphics[width=1.0\linewidth]{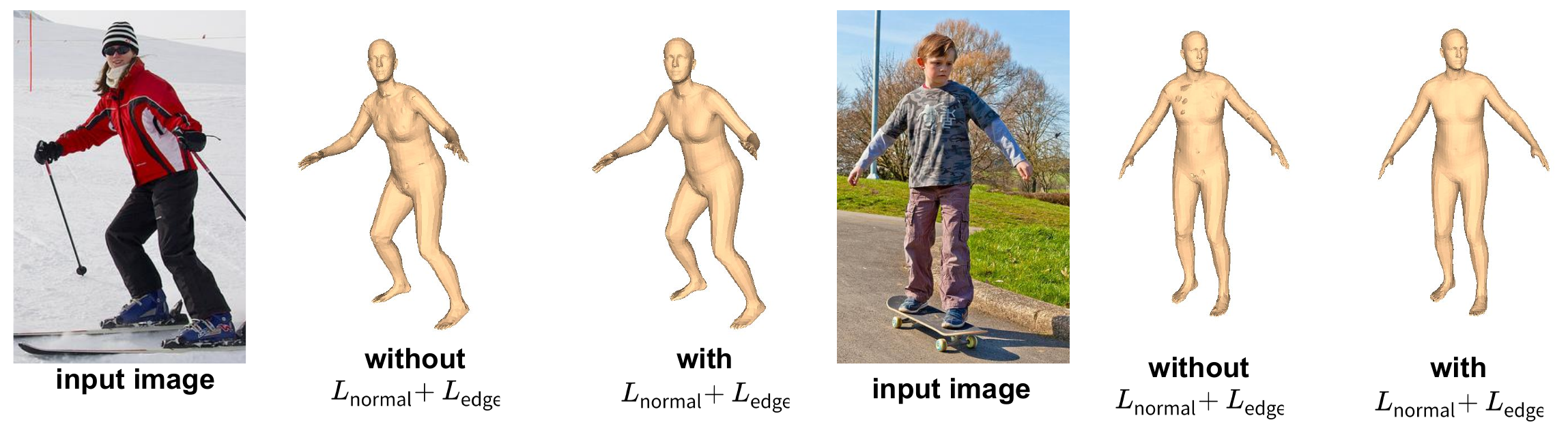}
\end{center}
\vspace*{-7mm}
   \caption{
Estimated meshes from models trained with different combinations of loss functions.
   }
\vspace*{-8mm}
\label{fig:normal_edge_loss}
\end{figure}

For visually pleasant mesh estimation, we use normal vector loss $L_\text{normal}$ and edge length loss $L_\text{edge}$.
We show the effectiveness of the two loss functions in Figure~\ref{fig:normal_edge_loss}.
As the figure shows, the two loss functions improves visual quality of output meshes.
We checked that $L_\text{normal}$ and $L_\text{edge}$ marginally affect the MPJPE and PA MPJPE.
For all settings, all models are trained on Human3.6M dataset and MSCOCO dataset.

\begin{table}
\centering
\setlength\tabcolsep{1.0pt}
\def\arraystretch{1.1}
\scalebox{1.0}{
\begin{tabular}{C{3.5cm}|C{2.0cm}C{2.0cm}}
\specialrule{.1em}{.05em}{.05em}
settings & MPJPE & PA MPJPE \\ \hline
wo. $L_\text{pose}^\text{MeshNet}$ & 84.5 & 58.5  \\
w. $L_\text{pose}^\text{MeshNet}$ & 81.8 & 58.0  \\ \hline
 \specialrule{.1em}{.05em}{.05em}
\end{tabular}
}
\caption{The MPJPE and PA MPJPE comparison between models trained with and without $L_\text{pose}^\text{MeshNet}$ on Human3.6M dataset.}
\vspace*{-7mm}
\label{table:meshnet_pose_loss}
\end{table}

\vspace*{-10mm}

\section{Accuracy of PoseNet}
We provide the MPJPE and PA MPJPE of PoseNet from I2L-MeshNet in Table~\ref{table:posenet_error}.
The PoseNet is trained with MeshNet by minimizing the loss function $L$.
As our PoseNet predicts 3D joint coordinates of the SMPL body joint set or MANO hand joint set, we calculate the errors using groundtruth SMPL or MANO 3D joint coordinates.
We could not calculate the MPJPE on FreiHAND dataset because the official evaluation server does not support it.

\begin{table}
\centering
\setlength\tabcolsep{1.0pt}
\def\arraystretch{1.1}
\scalebox{0.8}{
\begin{tabular}{C{3.5cm}|C{2.0cm}C{2.0cm}}
\specialrule{.1em}{.05em}{.05em}
datasets & MPJPE & PA MPJPE \\ \hline
Human3.6M & 62.2 & 47.2  \\
3DPW & 112.2 & 72.3  \\ 
SURREAL & 40.0 & 29.5  \\
FreiHAND & n/a & 8.0 \\ \hline
 \specialrule{.1em}{.05em}{.05em}
\end{tabular}
}
\caption{The MPJPE and PA MPJPE of PoseNet on each dataset.}
\vspace*{-7mm}
\label{table:posenet_error}
\end{table}

\newpage

\section{Pseudo-groundtruth SMPL parameters of Human3.6M dataset}
All the previous works~\cite{kanazawa2018end,pavlakos2018learning,kolotouros2019convolutional,kolotouros2019learning} used SMPL parameters obtained by applying Mosh~\cite{loper2014mosh} on the marker data of Human3.6M dataset as the groundtruth parameters.
However, currently, the distribution of the SMPL parameters from Mosh is disallowed because of the license problem.
In addition, the source code of Mosh is not publicly released.
Alternatively, we obtain groundtruth SMPL parameters by applying SMPLify-X~\cite{pavlakos2019expressive} on the groundtruth 3D joint coordinates of Human3.6M dataset.
Although the obtained SMPL parameters are not perfectly aligned to the groundtruth 3D joint coordinates, we checked that the error of the SMPLify-X is much less than those of current state-of-the-art 3D human pose estimation methods, as shown in Table~\ref{table:smplify-x_error}. 
Therefore, we think using SMPL parameters from SMPLify-X as groundtruth is reasonable.
Note that for a fair comparison, all the experimental results of previous works are reported by training and testing them on our SMPL parameters from SMPLify-X.
When fitting, we used neutral gender SMPL body model.
However, we found that it produces gender-specific body shapes, although we did not specify gender for each subject.
As most of the subjects of the training set in Human3.6M dataset are female, we found that our I2L-MeshNet trained on Human3.6M dataset tends to produce female body shape meshes.
We tried to fix the identity code of the SMPL body model obtained from the T-pose; however it produces higher errors.
Thus, we did not fix the identity code for each subject.

\begin{table}
\centering
\setlength\tabcolsep{1.0pt}
\def\arraystretch{1.1}
\scalebox{1.0}{
\begin{tabular}{C{4.5cm}|C{2.0cm}C{2.0cm}}
\specialrule{.1em}{.05em}{.05em}
methods & MPJPE  \\ \hline
Moon~et al.~\cite{moon2019camera} & 53.3  \\ 
Sun~et al.~\cite{sun2018integral} & 49.6  \\
Iskakov~et al.~\cite{iskakov2019learnable}*  & 20.8 \\
SMPLify-X from GT 3D pose & \textbf{13.1}  \\ \hline
 \specialrule{.1em}{.05em}{.05em}
\end{tabular}
}
\caption{The MPJPE comparison between SMPLify-X fitting results and state-of-the-art 3D human pose estimation methods. \enquote{*} takes multi-view RGB images as inputs.}
\vspace*{-7mm}
\label{table:smplify-x_error}
\end{table}

\newpage

\begin{table}[t]
\centering
\setlength\tabcolsep{1.0pt}
\def\arraystretch{1.1}
\begin{tabular}{C{3.5cm}|C{1.3cm}C{1.2cm}}
\specialrule{.1em}{.05em}{.05em}
methods & MPVPE & MPJPE \\ \hline
SMPLify~\cite{bogo2016keep} & 75.3 & - \\
BodyNet~\cite{varol2018bodynet} & 65.8 & 40.8 \\
\textbf{I2L-MeshNet (Ours)} & \textbf{44.7} & \textbf{37.7} \\
 \specialrule{.1em}{.05em}{.05em}
\end{tabular}
\caption{The MPVPE and MPJPE comparison between state-of-the-art methods and the proposed I2L-MeshNet on SURREAL.}
\vspace*{-5mm}
\label{table:compare_surreal}
\end{table}

\section{Evaluation on SURREAL}
We additionally provide evaluation results on SURREAL~\cite{varol17_surreal} that contains 67K clips synthesized by animating SMPL body model. 
We followed the same training and test set split of BodyNet~\cite{varol2018bodynet}.
For evaluation, mean per-vertex position error (MPVPE), which is averaged per-vertex Euclidean distance error (mm) between predicted and groundtruth 3D mesh coordinates, and MPJPE are used after root joint alignment.
We compare MPVPE and MPJPE of our I2L-MeshNet with previous state-of-the-art 3D human body pose and mesh estimation methods~\cite{bogo2016keep,kanazawa2018end,varol2018bodynet} on the SURREAL test set.
To this end, we reduced the clips in the training set to 1 fps to make the training image set.
Table~\ref{table:compare_surreal} shows that the proposed I2L-MeshNet significantly outperforms all previous state-of-the-art methods.
Especially, it achieves much lower test error compared with BodyNet~\cite{varol2018bodynet}, model-free approach.

\section{Qualitative results}
\begin{figure}
\begin{center}
\includegraphics[width=1.0\linewidth]{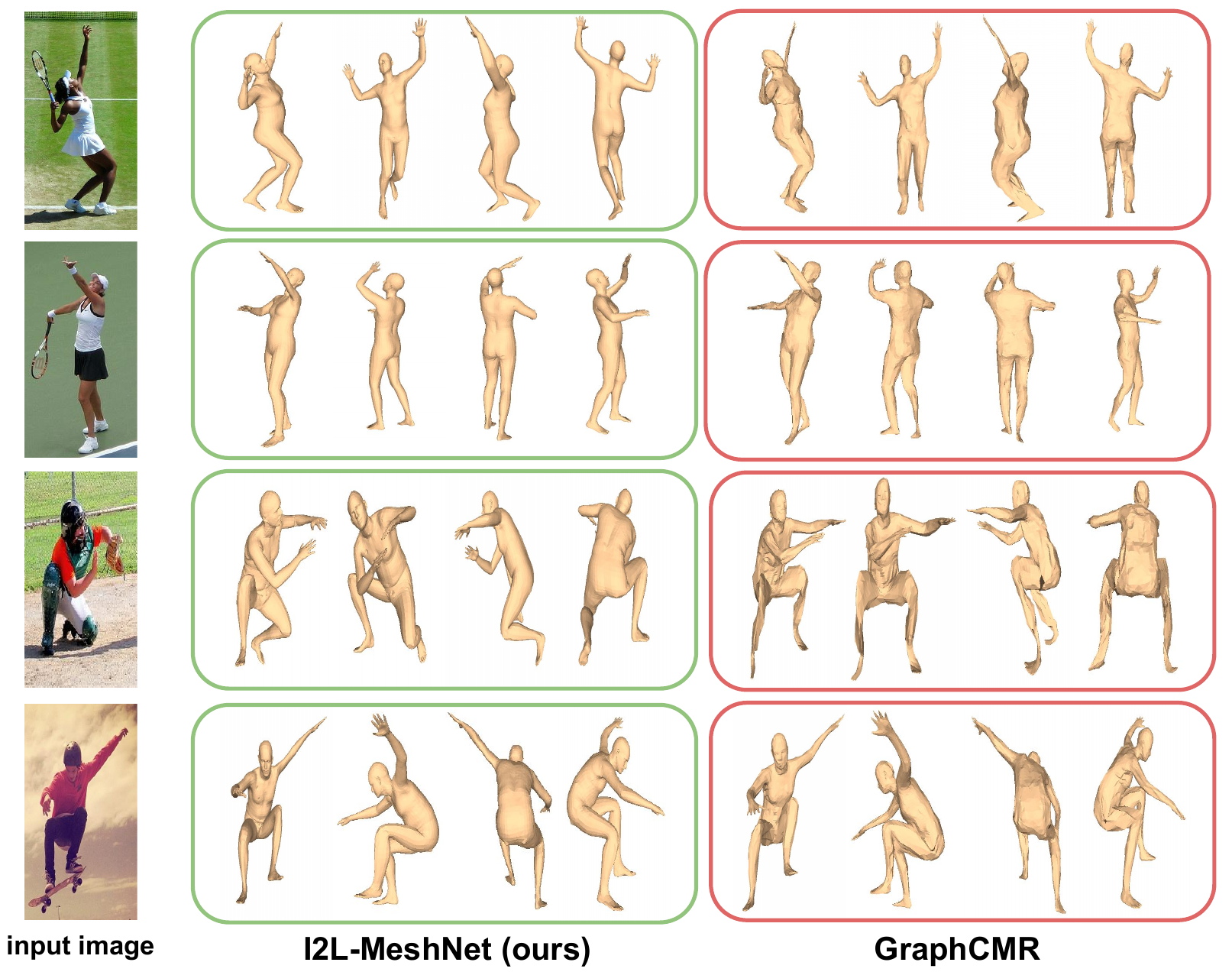}
\end{center}
\vspace*{-7mm}
   \caption{
    Estimated meshes comparisons between our I2L-MeshNet and GraphCMR~\cite{kolotouros2019convolutional}.
   }
\vspace*{-8mm}
\label{fig:qualitative}
\end{figure}

We provide qualitative results comparison between ours and previous state-of-the-art model-free method (\textit{i.e.}, GraphCMR~\cite{kolotouros2019convolutional}) in Figure~\ref{fig:qualitative}.
As the figure shows, our I2L-MeshNet provides much more visually pleasant mesh results than GraphCMR.
We think this is because the graph convolutional network (GraphCNN) often tends to smooth the meshes by averaging the vertex feature with that of neighboring vertices.

\clearpage

%% file: main.bbl
\begin{thebibliography}{10}
\providecommand{\url}[1]{\texttt{#1}}
\providecommand{\urlprefix}{URL }
\providecommand{\doi}[1]{https://doi.org/#1}

\bibitem{andriluka20142d}
Andriluka, M., Pishchulin, L., Gehler, P., Schiele, B.: {2D} human pose
  estimation: New benchmark and state of the art analysis. In: CVPR (2014)

\bibitem{arnab2019exploiting}
Arnab, A., Doersch, C., Zisserman, A.: Exploiting temporal context for {3D}
  human pose estimation in the wild. In: CVPR (2019)

\bibitem{baek2019pushing}
Baek, S., In~Kim, K., Kim, T.K.: Pushing the envelope for {RGB}-based dense
  {3D} hand pose estimation via neural rendering. In: CVPR (2019)

\bibitem{bogo2016keep}
Bogo, F., Kanazawa, A., Lassner, C., Gehler, P., Romero, J., Black, M.J.: {Keep
  it SMPL}: Automatic estimation of {3D} human pose and shape from a single
  image. In: ECCV (2016)

\bibitem{boukhayma20193d}
Boukhayma, A., de~Bem, R., Torr, P.H.: {3D} hand shape and pose from images in
  the wild. In: CVPR (2019)

\bibitem{chen2018cascaded}
Chen, Y., Wang, Z., Peng, Y., Zhang, Z., Yu, G., Sun, J.: Cascaded pyramid
  network for multi-person pose estimation. In: CVPR (2018)

\bibitem{choi2020p2m}
Choi, H., Moon, G., Lee, K.M.: {Pose2Mesh}: Graph convolutional network for
  {3D} human pose and mesh recovery from a {2D} human pose. ECCV  (2020)

\bibitem{Freihand2019}
Christian~Zimmermann, Duygu~Ceylan, J.Y.B.R.M.A., Brox, T.: {FreiHAND}: A
  dataset for markerless capture of hand pose and shape from single {RGB}
  images. In: ICCV (2019)

\bibitem{ge20193d}
Ge, L., Ren, Z., Li, Y., Xue, Z., Wang, Y., Cai, J., Yuan, J.: {3D} hand shape
  and pose estimation from a single {RGB} image. In: CVPR (2019)

\bibitem{hasson2019learning}
Hasson, Y., Varol, G., Tzionas, D., Kalevatykh, I., Black, M.J., Laptev, I.,
  Schmid, C.: Learning joint reconstruction of hands and manipulated objects.
  In: CVPR (2019)

\bibitem{he2017mask}
He, K., Gkioxari, G., Doll{\'a}r, P., Girshick, R.: Mask r-cnn. In: ICCV (2017)

\bibitem{he2016deep}
He, K., Zhang, X., Ren, S., Sun, J.: Deep residual learning for image
  recognition. In: CVPR (2016)

\bibitem{ioffe2015batch}
Ioffe, S., Szegedy, C.: {Batch Normalization}: Accelerating deep network
  training by reducing internal covariate shift. ICML  (2015)

\bibitem{ionescu2014human3}
Ionescu, C., Papava, D., Olaru, V., Sminchisescu, C.: {Human3.6M}: Large scale
  datasets and predictive methods for {3D} human sensing in natural
  environments. TPAMI  (2014)

\bibitem{iqbal2018hand}
Iqbal, U., Molchanov, P., Breuel Juergen~Gall, T., Kautz, J.: Hand pose
  estimation via latent 2.5 d heatmap regression. In: ECCV (2018)

\bibitem{iskakov2019learnable}
Iskakov, K., Burkov, E., Lempitsky, V., Malkov, Y.: Learnable triangulation of
  human pose. In: ICCV (2019)

\bibitem{johnson2010clustered}
Johnson, S., Everingham, M.: Clustered pose and nonlinear appearance models for
  human pose estimation. In: BMVC (2010)

\bibitem{johnson2011learning}
Johnson, S., Everingham, M.: Learning effective human pose estimation from
  inaccurate annotation. In: CVPR (2011)

\bibitem{joo2020exemplar}
Joo, H., Neverova, N., Vedaldi, A.: Exemplar fine-tuning for 3d human pose
  fitting towards in-the-wild 3d human pose estimation. arXiv preprint
  arXiv:2004.03686  (2020)

\bibitem{kanazawa2018end}
Kanazawa, A., Black, M.J., Jacobs, D.W., Malik, J.: End-to-end recovery of
  human shape and pose. In: CVPR (2018)

\bibitem{kanazawa2019learning}
Kanazawa, A., Zhang, J.Y., Felsen, P., Malik, J.: Learning {3D} human dynamics
  from video. In: CVPR (2019)

\bibitem{kingma2014adam}
Kingma, D.P., Ba, J.: Adam: A method for stochastic optimization. ICLR  (2014)

\bibitem{kolotouros2019learning}
Kolotouros, N., Pavlakos, G., Black, M.J., Daniilidis, K.: Learning to
  reconstruct {3D} human pose and shape via model-fitting in the loop. In: ICCV
  (2019)

\bibitem{kolotouros2019convolutional}
Kolotouros, N., Pavlakos, G., Daniilidis, K.: Convolutional mesh regression for
  single-image human shape reconstruction. In: CVPR (2019)

\bibitem{lassner2017unite}
Lassner, C., Romero, J., Kiefel, M., Bogo, F., Black, M.J., Gehler, P.V.: Unite
  the people: Closing the loop between {3D} and {2D} human representations. In:
  CVPR (2017)

\bibitem{lin2014microsoft}
Lin, T.Y., Maire, M., Belongie, S., Hays, J., Perona, P., Ramanan, D.,
  Doll{\'a}r, P., Zitnick, C.L.: Microsoft coco: Common objects in context. In:
  ECCV (2014)

\bibitem{loper2014mosh}
Loper, M., Mahmood, N., Black, M.J.: Mosh: Motion and shape capture from sparse
  markers. ACM TOG  (2014)

\bibitem{loper2015smpl}
Loper, M., Mahmood, N., Romero, J., Pons-Moll, G., Black, M.J.: {SMPL}: A
  skinned multi-person linear model. ACM TOG  (2015)

\bibitem{von2018recovering}
von Marcard, T., Henschel, R., Black, M.J., Rosenhahn, B., Pons-Moll, G.:
  Recovering accurate {3D} human pose in the wild using imus and a moving
  camera. In: ECCV (2018)

\bibitem{mehta2017monocular}
Mehta, D., Rhodin, H., Casas, D., Fua, P., Sotnychenko, O., Xu, W., Theobalt,
  C.: Monocular {3D} human pose estimation in the wild using improved cnn
  supervision. In: 3DV (2017)

\bibitem{mehta2018single}
Mehta, D., Sotnychenko, O., Mueller, F., Xu, W., Sridhar, S., Pons-Moll, G.,
  Theobalt, C.: Single-shot multi-person {3D} pose estimation from monocular
  {RGB}. In: 3DV (2018)

\bibitem{moon2018v2v}
Moon, G., Chang, J.Y., Lee, K.M.: {V2V-PoseNet}: Voxel-to-voxel prediction
  network for accurate {3D} hand and human pose estimation from a single depth
  map. In: CVPR (2018)

\bibitem{moon2019camera}
Moon, G., Chang, J.Y., Lee, K.M.: Camera distance-aware top-down approach for
  {3D} multi-person pose estimation from a single {RGB} image. In: ICCV (2019)

\bibitem{newell2016stacked}
Newell, A., Yang, K., Deng, J.: Stacked hourglass networks for human pose
  estimation. In: ECCV (2016)

\bibitem{omran2018neural}
Omran, M., Lassner, C., Pons-Moll, G., Gehler, P., Schiele, B.: {Neural Body
  Fitting}: Unifying deep learning and model based human pose and shape
  estimation. In: 3DV. IEEE (2018)

\bibitem{paszke2017automatic}
Paszke, A., Gross, S., Chintala, S., Chanan, G., Yang, E., DeVito, Z., Lin, Z.,
  Desmaison, A., Antiga, L., Lerer, A.: Automatic differentiation in pytorch
  (2017)

\bibitem{pavlakos2019expressive}
Pavlakos, G., Choutas, V., Ghorbani, N., Bolkart, T., Osman, A.A., Tzionas, D.,
  Black, M.J.: Expressive body capture: {3D} hands, face, and body from a
  single image. In: CVPR (2019)

\bibitem{pavlakos2019texturepose}
Pavlakos, G., Kolotouros, N., Daniilidis, K.: {TexturePose}: Supervising human
  mesh estimation with texture consistency. In: ICCV (2019)

\bibitem{pavlakos2017coarse}
Pavlakos, G., Zhou, X., Derpanis, K.G., Daniilidis, K.: Coarse-to-fine
  volumetric prediction for single-image {3D} human pose. In: CVPR (2017)

\bibitem{pavlakos2018learning}
Pavlakos, G., Zhu, L., Zhou, X., Daniilidis, K.: Learning to estimate {3D}
  human pose and shape from a single color image. In: CVPR (2018)

\bibitem{rogez2017lcr}
Rogez, G., Weinzaepfel, P., Schmid, C.: {LCR-Net}:
  Localization-classification-regression for human pose. In: CVPR (2017)

\bibitem{romero2017embodied}
Romero, J., Tzionas, D., Black, M.J.: Embodied hands: Modeling and capturing
  hands and bodies together. ACM TOG  (2017)

\bibitem{russakovsky2015imagenet}
Russakovsky, O., Deng, J., Su, H., Krause, J., Satheesh, S., Ma, S., Huang, Z.,
  Karpathy, A., Khosla, A., Bernstein, M., et~al.: Imagenet large scale visual
  recognition challenge. IJCV  (2015)

\bibitem{sun2018integral}
Sun, X., Xiao, B., Wei, F., Liang, S., Wei, Y.: Integral human pose regression.
  In: ECCV (2018)

\bibitem{tompson2014joint}
Tompson, J.J., Jain, A., LeCun, Y., Bregler, C.: Joint training of a
  convolutional network and a graphical model for human pose estimation. In:
  NeurIPS (2014)

\bibitem{varol2018bodynet}
Varol, G., Ceylan, D., Russell, B., Yang, J., Yumer, E., Laptev, I., Schmid,
  C.: {BodyNet}: Volumetric inference of {3D} human body shapes. In: ECCV
  (2018)

\bibitem{varol17_surreal}
Varol, G., Romero, J., Martin, X., Mahmood, N., Black, M.J., Laptev, I.,
  Schmid, C.: Learning from synthetic humans. In: CVPR (2017)

\bibitem{wang2018pixel2mesh}
Wang, N., Zhang, Y., Li, Z., Fu, Y., Liu, W., Jiang, Y.G.: {Pixel2Mesh}:
  Generating {3D} mesh models from single {RGB} images. In: ECCV (2018)

\bibitem{xiao2018simple}
Xiao, B., Wu, H., Wei, Y.: Simple baselines for human pose estimation and
  tracking. In: ECCV (2018)

\bibitem{xu2019denserac}
Xu, Y., Zhu, S.C., Tung, T.: {DenseRaC}: Joint {3D} pose and shape estimation
  by dense render-and-compare. In: ICCV (2019)

\end{thebibliography}
